\documentclass[letterpaper, 10 pt, conference]{ieeeconf}  

\IEEEoverridecommandlockouts                              

\usepackage[utf8]{inputenc}
\usepackage{graphics} 
\usepackage{graphicx}
\usepackage{epsfig} 
\usepackage{mathptmx} 
\usepackage{times} 
\usepackage{amsmath} 
\usepackage{amssymb}  
\usepackage{subcaption} 
\usepackage{url}
\usepackage{hyperref}

\title{\LARGE \bf
Direct Data-Driven Predictive Control for a Three-Dimensional Cable-Driven Soft Robotic Arm}

\author{Cheng Ouyang$^{1}$, Moeen Ul Islam$^{1}$, Dong Chen$^{1}$, Kaixiang Zhang$^{2}$, Zhaojian Li$^{2}$, and Xiaobo Tan$^{3}$
\thanks{$^{1}$Cheng Ouyang, Moeen Ul Islam, and Dong Chen are with the Department of Agricultural \& Biological Engineering, Mississippi State University, MS, USA. {\tt\small Emails: \{co603, mu136, dc2528\}@msstate.edu}}%
\thanks{$^{2}$Kaixiang Zhang and Zhaojian Li are with the Department of Mechanical Engineering, Michigan State University, MI, USA.
        {\tt\small Emails: \{zhangk64, lizhaoj1\}@msu.edu}}%
\thanks{$^{3}$Xiaobo Tan is with the Department of Electrical and Computer Engineering, Michigan State University, MI, USA.
{\tt\small Email: xbtan@egr.msu.edu}}%
}

\begin{document}

\maketitle
\thispagestyle{empty}
\pagestyle{empty}

\begin{abstract}
Soft robots offer significant advantages in safety and adaptability, yet achieving precise and dynamic control remains a major challenge due to their inherently complex and nonlinear dynamics. Recently, \underline{D}ata-\underline{e}nabl\underline{e}d \underline{P}redictive \underline{C}ontrol (DeePC) has emerged as a promising model-free method that bypasses explicit system identification by directly leveraging input–output data. While DeePC has shown success in other domains, its application to soft robots remains underexplored, particularly for 3D soft robotic systems. This paper addresses this gap by developing and experimentally validating an effective DeePC framework on a 3D cable-driven soft arm. Specifically, we design and fabricate a soft robotic arm with a thick tubing backbone for stability, a dense silicone body with large cavities for strength and flexibility, and rigid endcaps for secure termination. Using this platform, we implement DeePC with singular value decomposition (SVD)-based dimension reduction for two key control tasks: fixed-point regulation and trajectory tracking in 3D space. Comparative experiments with a baseline model-based controller demonstrate DeePC’s superior accuracy and adaptability, highlighting its potential as a practical solution for dynamic control of soft robots.
\end{abstract}

\section{INTRODUCTION}
Soft robots, enabled by compliant materials and novel morphologies, have attracted increasing research interest across diverse applications, such as fruit harvesting \cite{wang2023development}, medical procedures \cite{cianchetti2018biomedical}, and beyond \cite{lee2017soft}. Compared to conventional rigid robots, they offer safer human–robot interaction \cite{bao2018soft, das2019review}, greater adaptability to unstructured environments \cite{chen2025survey}, and theoretically unlimited degrees of freedom for versatile manipulation \cite{rus2015design}. 

However, this promise comes with profound control challenges. The very sources of soft robots’ advantages, deformable materials and continuum structures, also lead to extreme complexity in modeling and control \cite{wang2022control}. A primary difficulty lies in their highly nonlinear dynamics: materials such as silicone exhibit nonlinear stress–strain behavior, and under pneumatic, electrical, or tendon actuation, their responses often involve strong nonlinearity, hysteresis, and creep \cite{della2023model}. Another challenge is the infinite-dimensional state space. Unlike rigid robots, whose configurations can be described by a finite set of joint angles and velocities, a continuously deformable body theoretically requires specifying the displacement of every point, resulting in an infinite-dimensional system. Collectively, these factors make soft robot control particularly challenging \cite{zheng2025estimating}.

To address the complex dynamics of soft robots, researchers have proposed numerous model-based control strategies; however, these methods often struggle to balance accuracy and computational complexity \cite{della2023model}. For instance, the Piecewise Constant Curvature (PCC) model is the most widely used simplification \cite{katzschmann2019dynamic}. It assumes each section bends as a circular arc with constant curvature, reducing the configuration to only a few parameters. Yet, PCC’s validity depends on this assumption, which is frequently violated in practice due to gravity, environmental interactions, and non-uniform actuation or materials, resulting in significant model–plant mismatch and degraded performance \cite{roshanfar2023hyperelastic}. The finite element method (FEM) offers a more accurate alternative, capable of capturing complex geometries, nonlinear material properties, and contact interactions between soft robots and their environments. However, this accuracy comes at the cost of high computational demands \cite{chenevier2018reduced}. To address this, Navez et al. proposed a condensed FEM model that combines the efficiency of machine learning with the physical interpretability of FEM-based mechanics \cite{navez2025modeling}. While promising, it relies on quasi-static assumptions, requires extensive FEM-generated training data, and shows limited generalization to highly nonlinear or dynamic scenarios.

Given the limitations of first-principles modeling, data-driven methods provide a promising alternative for soft robot control. Instead of deriving exact physical models, these approaches learn input–output mappings or latent dynamics directly from experimental data. Reinforcement learning (RL) offers an appealing model-free control strategy for systems that are difficult to model; however, its use is hindered by low sample efficiency, the simulation-to-reality gap, and safety concerns associated with exploration \cite{qi2024back, garg2025autonomous}. Koopman operator theory provides a distinct, data-driven approach. Unlike RL, it is not model-free; rather, it identifies an explicit linear model from data that represents the dynamics of the underlying nonlinear system. Bruder et al. \cite{bruder2020data} demonstrated a Koopman-based method that enabled real-time trajectory tracking of soft robots with high computational efficiency. However, its effectiveness relies heavily on the choice of basis functions, and for more complex systems, the model dimension can become computationally intractable. 

In contrast to RL’s fully model-free nature and Koopman operator theory’s explicit modeling, data-enabled predictive control (DeePC) adopts a nonparametric, model-implicit approach by leveraging the Hankel matrix of past input–output data as its predictive model \cite{coulson2019data}. DeePC has been successfully applied in several domains, including large-scale damping control in power systems \cite{huang2021decentralized}, building automation \cite{chinde2022data}, quadrotor trajectory tracking \cite{elokda2021data}, and power-converter control \cite{huang2021quadratic}. Its robustness, ability to handle nonlinearity and uncertainty, and nonparametric nature align closely with the challenges of soft robot control \cite{huang2023robust}, yet related applications remain at an early stage. Wang et al. were the first to apply DeePC to soft robots \cite{wang2024mechanical}, but their validation was limited to a planar, pneumatically actuated system restricted to one plane of motion. Extending DeePC to three-dimensional soft arms and validating it in high-precision 3D dynamic tasks thus remains a key open problem. To address this gap, we build on our previous work on a modular cable-driven soft robotic arm, for which we developed a nonlinear kinematic model and demonstrated open-loop control \cite{qi2024design}. This study advances that foundation by (i) optimizing the robotic arm design, (ii) introducing a DeePC framework with singular value decomposition (SVD)-based dimension reduction, systematically implemented and experimentally validated on the developed soft robotic arm platform for two control tasks: fixed-point regulation and dynamic trajectory tracking in 3D space, and (iii) open-sourcing the arm designs and control codes to facilitate future research\footnote{Codes and designs: \url{https://github.com/chengoy30/DeePC-for-Soft-Robot-Control}}.

\section{Hardware Design}
This section presents the hardware design of the soft robot, including its structural configuration, fabrication process, and control modules.

\subsection{Robot design and fabrication}
\begin{figure}[htbp]
    \centering
    \begin{subfigure}{0.2\textwidth}
        \centering
        \includegraphics[width=\linewidth]{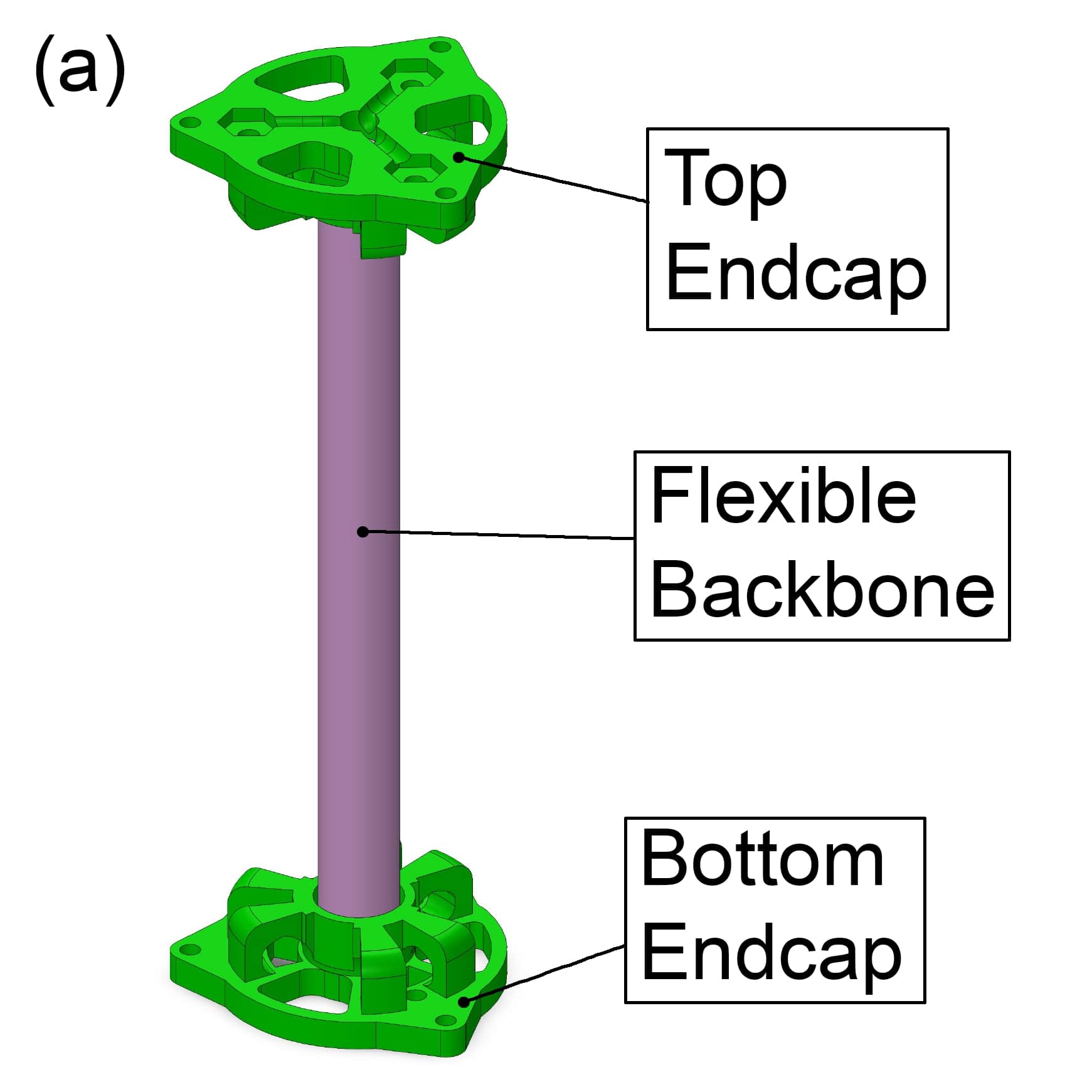}
        \phantomcaption
        \label{fig:softrobot_a}
    \end{subfigure}
    \begin{subfigure}{0.2\textwidth}
        \centering
        \includegraphics[width=\linewidth]{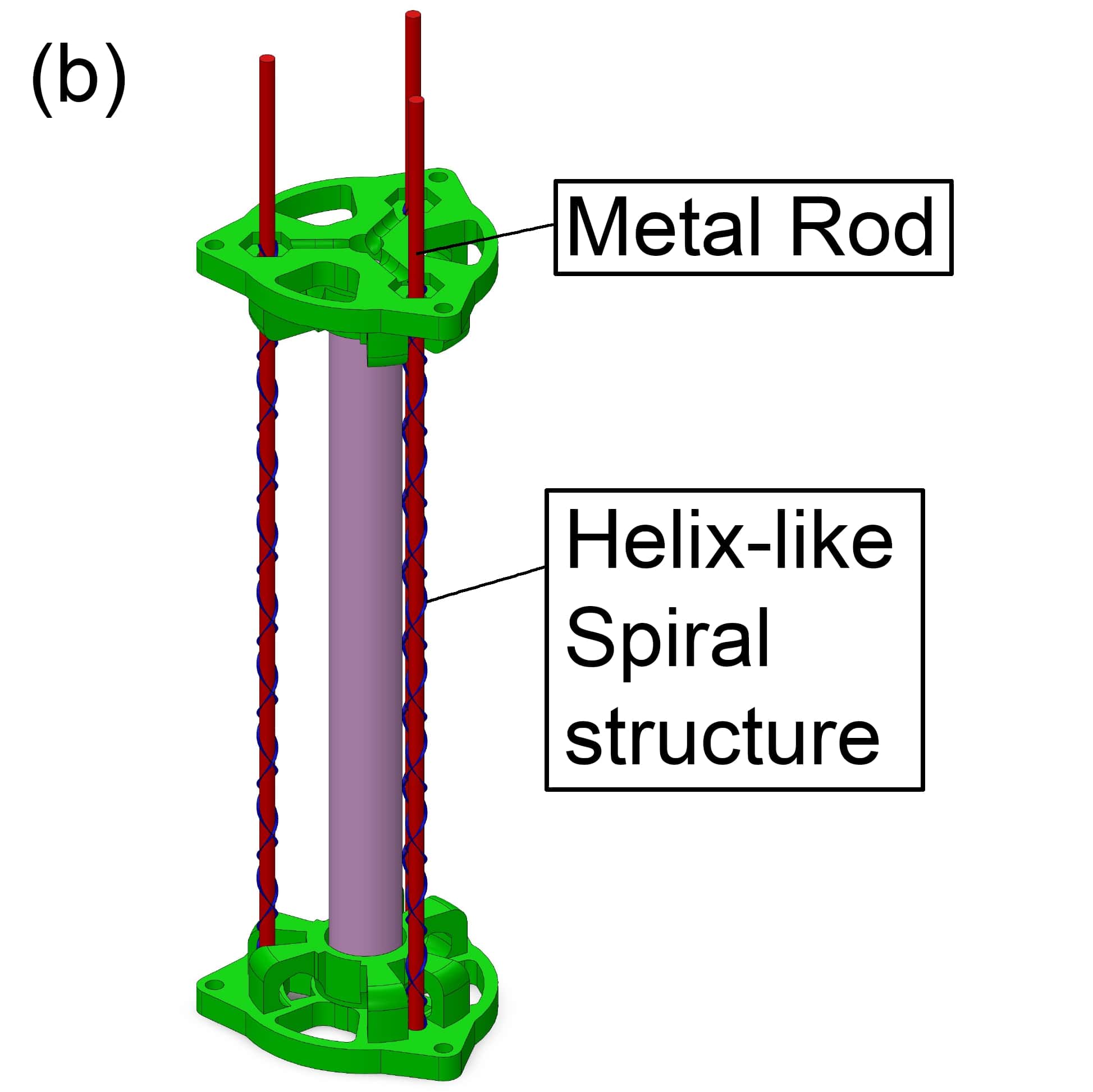}
        \phantomcaption
        \label{fig:softrobot_b}
    \end{subfigure}
    \vspace{1.5em}
    \begin{subfigure}{0.2\textwidth}
        \centering
        \includegraphics[width=\linewidth]{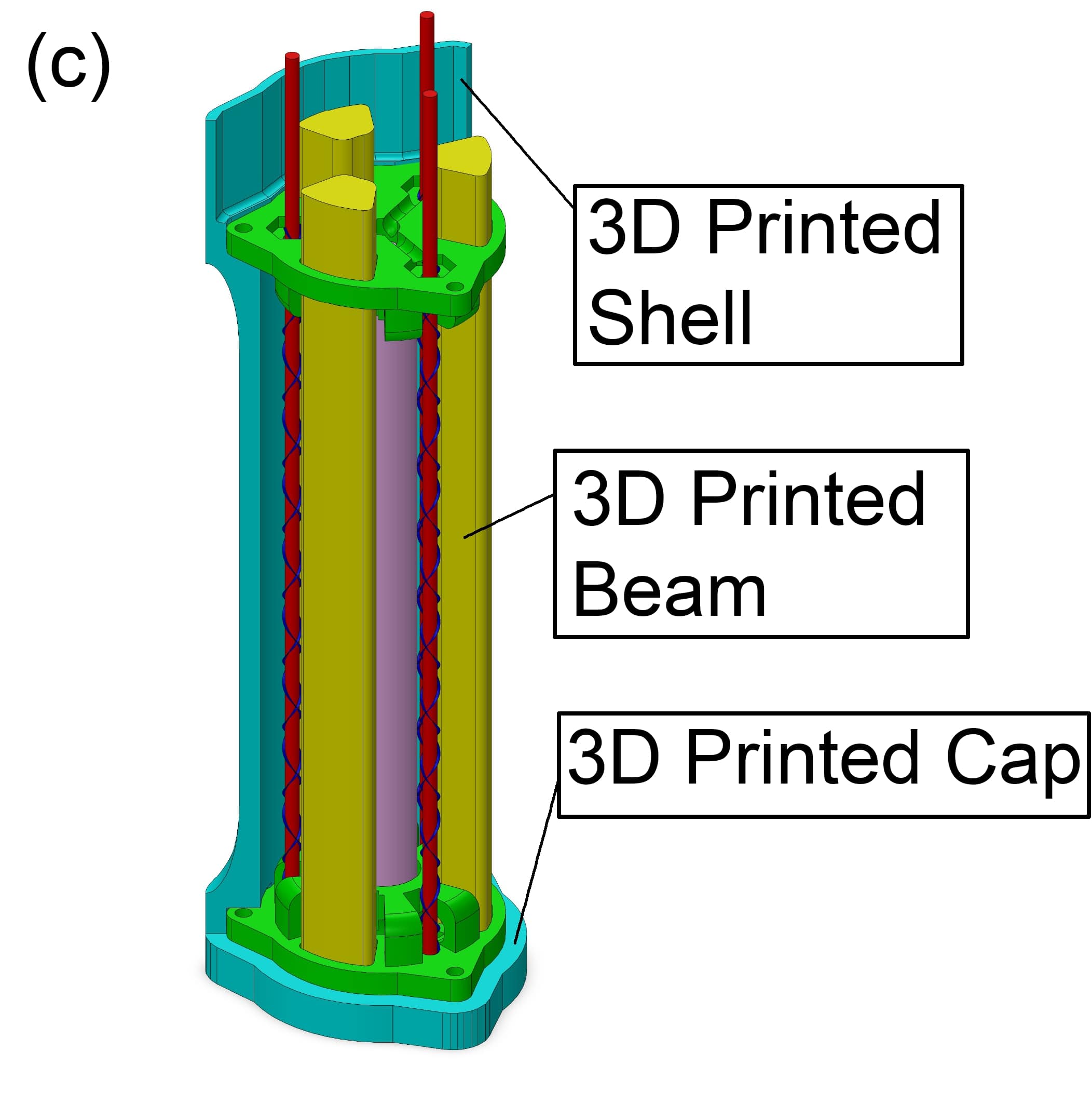}
        \phantomcaption
        \label{fig:softrobot_c}
    \end{subfigure}
    \begin{subfigure}{0.2\textwidth}
        \centering
        \includegraphics[width=\linewidth]{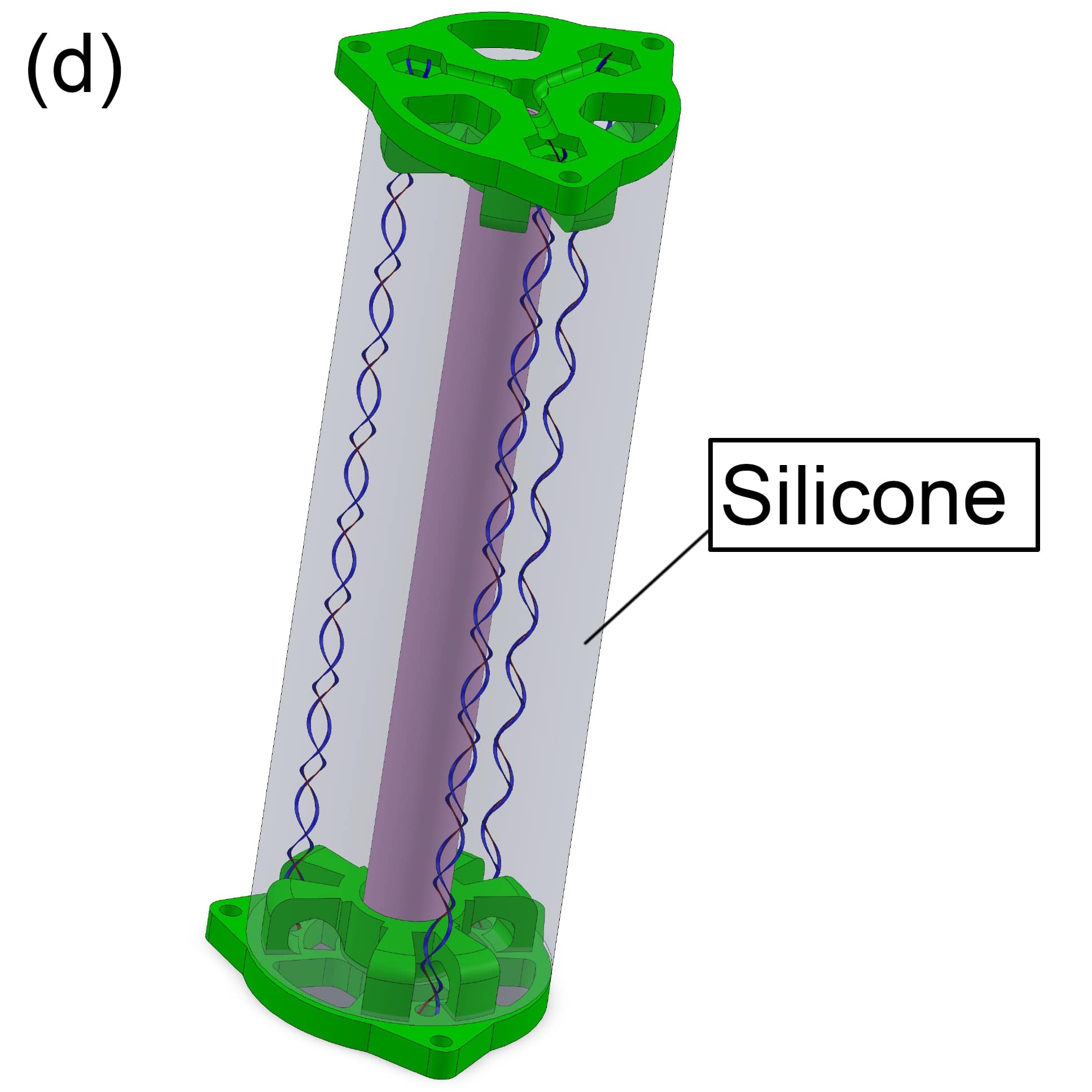}
        \phantomcaption
        \label{fig:softrobot_d}
    \end{subfigure}
    \vspace{-20pt}
    \caption{Fabrication process of the soft robot: (a–c) sequential setup steps prior to silicone casting, and (d) the completed soft robot after curing. Each panel is labeled to indicate the corresponding components.}
    \label{fig:softrobot_full}
    \vspace{-12pt}
\end{figure}

To apply DeePC, we initially tested our previous robot design in \cite{qi2024design}. However, through repeated testing and several design iterations, we found that practical applications require greater stiffness and load-bearing capacity while preserving the inherent flexibility of soft robots.
To address this, the final design employed a thicker flexible backbone and denser silicone, providing the necessary structural strength while maintaining compliance. The resulting soft robotic arm consists of two end caps, a flexible backbone, and three helix-like spiral structures that incorporate Kevlar threads for cable routing. The detailed structure and sequential fabrication process are shown in Fig.~\ref{fig:softrobot_a}–\ref{fig:softrobot_c}, while the final cured soft robot is presented in Fig.~\ref{fig:softrobot_d}. The end caps (Fig.~\ref{fig:softrobot_a}) were 3D-printed using PETG, with anchor points designed to prevent detachment of the silicone after casting, and further shaped to support the mounting and stacking of multiple arm segments. A flexible backbone was formed using clear Masterkleer soft PVC tubing (McMaster-Carr) with an outer diameter of 8 mm and an inner diameter of 3 mm. The molds for silicone casting (Fig.~\ref{fig:softrobot_c}) were also fabricated via 3D printing, and Ecoflex 00-50 silicone was used as the elastomer material. High-strength Kevlar threads served dual purposes: as actuation cables and as reinforcement in the helix-like spiral structures.

\subsection{Actuation and control system}
\begin{figure}[htbp]
    \centering
    \begin{subfigure}{0.23\textwidth}
        \centering
        \includegraphics[width=\linewidth]
        {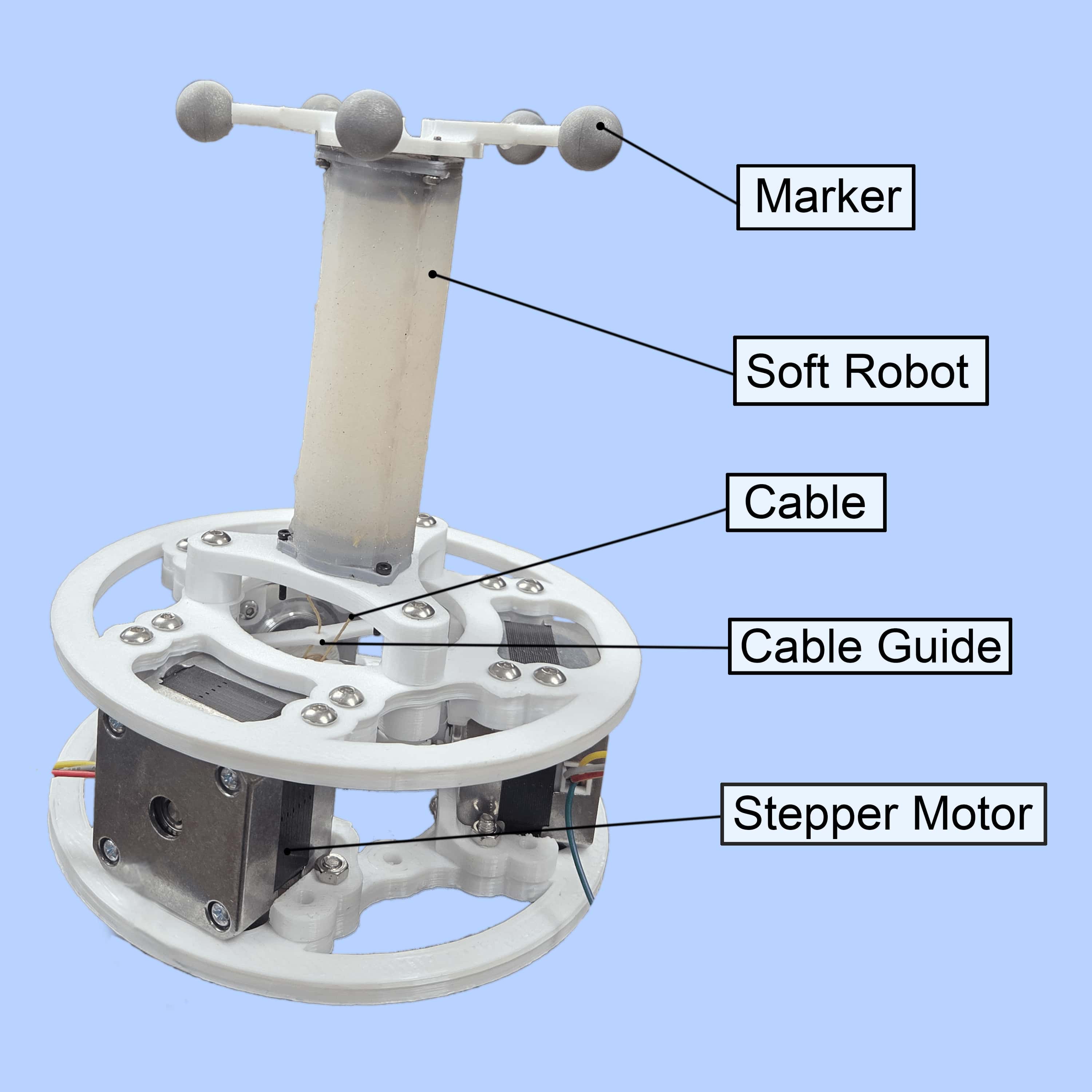}
        \caption{}
        \label{fig:hardware1}
    \end{subfigure}\hfill
    \begin{subfigure}{0.23\textwidth}
        \centering
        \includegraphics[width=\linewidth]{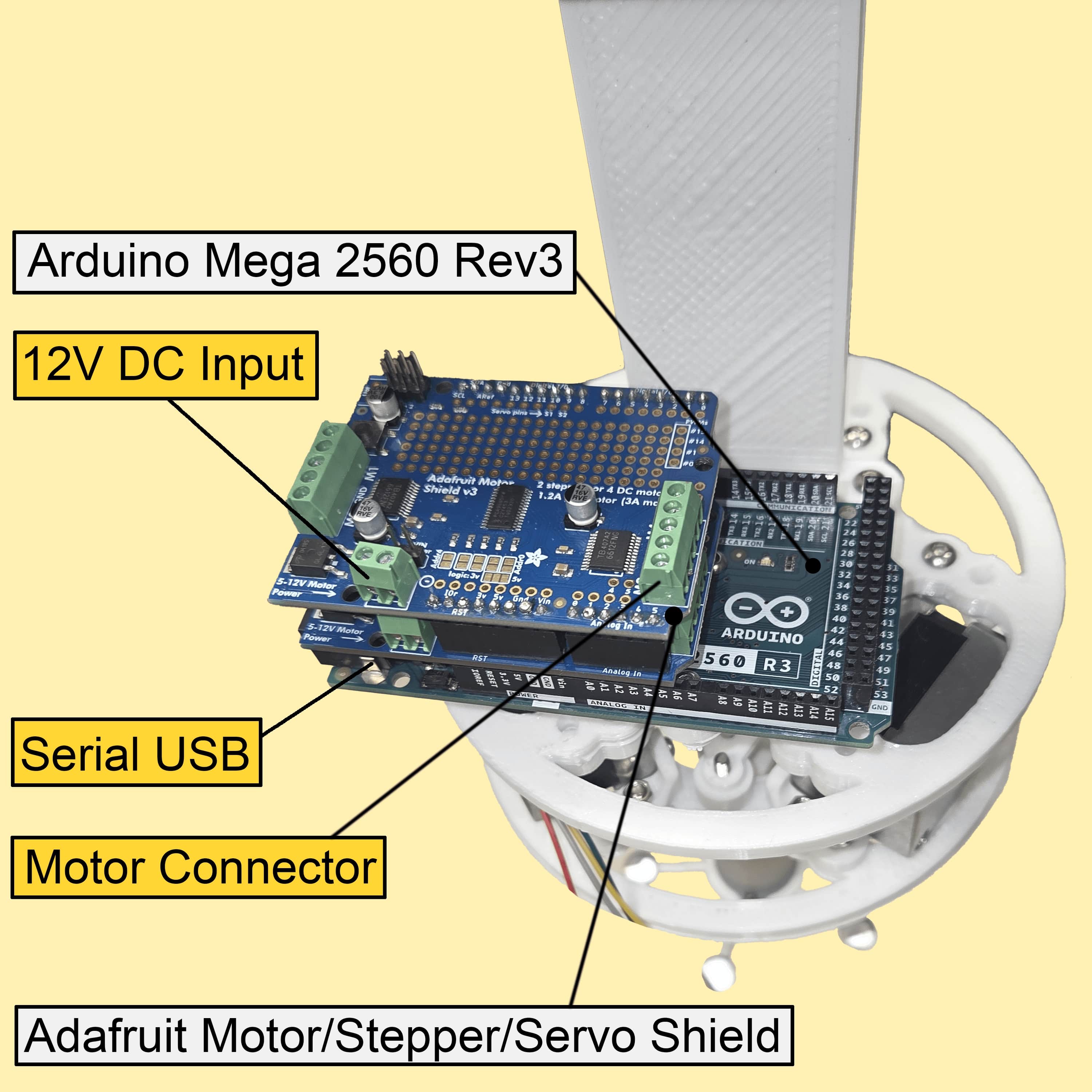}
        \caption{}
        \label{fig:hardware2}
    \end{subfigure}
    \caption{Hardware setup of the soft robot, illustrating (a) the actuation system and (b) the low-level control system.}
    \label{fig:hardware}
    \vspace{-14pt}
\end{figure}

The soft robot arm is cable-driven, with cable lengths precisely regulated by three stepper motors. Each actuation cable is driven by a 3D-printed pulley attached directly to a stepper motor. The three stepper motors are mounted on a 3D-printed frame at 120° separation. The frame also incorporates dedicated cable guides that route each actuation cable to its corresponding stepper motor. This design prevents tangling and ensures the cables remain properly seated within the pulleys. The overall mechanical structure of the system is shown in Fig.~\ref{fig:hardware1}.

For motor control, an Arduino Mega 2560 Rev3 serves as the low-level microcontroller. It drives the stepper motors via a motor driver shield (Adafruit Motor Shield V2) and operates in open-loop mode without encoder feedback. The complete control board is shown in Fig.~\ref{fig:hardware2}. The low-level controller receives commands from a high-level controller (a laptop) through serial communication, while the high-level controller computes the optimal control inputs and generates the corresponding commands.

\section{DeePC for soft robot control} \label{sec:deepc}
\subsection{Data-enabled Predictive Control (DeePC)}
We begin with an overview of DeePC \cite{coulson2019data}. By linearizing the system dynamics around a nominal operating point, the behavior of a soft robot can be approximated by a discrete-time linear time-invariant (LTI) system:
\begin{equation}
\begin{aligned}
x(t+1) &= A x(t) + B u(t), \\
y(t) &= C x(t) + D u(t),
\end{aligned}
\label{eq1}
\end{equation}
where $A \in \mathbb{R}^{n \times n}, B \in \mathbb{R}^{n \times m}, C \in \mathbb{R}^{p \times n}, D \in \mathbb{R}^{p \times m}$ are the system matrices, and $x(t) \in \mathbb{R}^n$, $u(t) \in \mathbb{R}^m$, and $y(t) \in \mathbb{R}^p$ denote the state, input, and output, respectively. This formulation is applicable to soft robotic systems whose locally linearized LTI models are controllable and observable.

The core idea of DeePC is to bypass explicit system identification by directly utilizing offline input–output data rather than a parametric model \cite{coulson2019data}. This approach is grounded in behavioral systems theory, which represents the system’s dynamics using a block Hankel matrix \cite{willems2005note}. For the specific case of our soft robotic arm, we begin by conducting an offline data collection experiment. A predefined sequence of control inputs $u^\mathrm{d} = \begin{bmatrix} u^{\mathrm{d}}(0)^{\top}, u^{\mathrm{d}}(1)^{\top}, \ldots, u^{\mathrm{d}}(T-1)^{\top}\end{bmatrix}^{\top} \in \mathbb{R}^{mT}$, which denotes the incremental lengths of the three actuation cables, is applied to the robotic arm. Simultaneously, the resulting trajectory of the end-effector is recorded as the output sequence $y^{\mathrm{d}} = \begin{bmatrix} y^{\mathrm{d}}(0)^{\top}, y^{\mathrm{d}}(1)^{\top}, \ldots, y^{\mathrm{d}}(T-1)^{\top} \end{bmatrix}^{\top} \in \mathbb{R}^{pT}$. This collected input-output dataset is then structured into a block Hankel matrix of depth $L\in \mathbb{Z}$. This data matrix serves as a non-parametric model of the robot's dynamics and is defined as:
\begin{equation}
\begin{bmatrix}
\mathcal{H}_{L}(u^\mathrm{d}) \\ \hline
\mathcal{H}_{L}(y^\mathrm{d})
\end{bmatrix}
=
\left[
  \begin{array}{cccc}
    u^\mathrm{d}(0) & u^\mathrm{d}(1) & \cdots & u^\mathrm{d}(T-L) \\
    \vdots & \vdots & \ddots & \vdots \\
    u^\mathrm{d}(L-1) & u^\mathrm{d}(L) & \cdots & u^\mathrm{d}(T-1) \\
    \hline
    y^\mathrm{d}(0) & y^\mathrm{d}(1) & \cdots & y^\mathrm{d}(T-L) \\
    \vdots & \vdots & \ddots & \vdots \\
    y^\mathrm{d}(L-1) & y^\mathrm{d}(L) & \cdots & y^\mathrm{d}(T-1)
  \end{array}
\right].
\end{equation}

\textit{Lemma 1 (Fundamental Lemma \cite{willems2005note})}:
Consider the controllable LTI system in \eqref{eq1}, and suppose that the input sequence $u^\mathrm{d}$ is persistently exciting of order $n+L$. Then, any length-$L$ input–output trajectory $(u_{[0,L-1]},y_{[0,L-1]})$ of \eqref{eq1} can be represented if and only if there exists a real vector $g \in \mathbb{R}^{T-L+1}$ such that
\begin{equation}
\begin{bmatrix}
u_{[0,L-1]} \\
y_{[0,L-1]}
\end{bmatrix}
=
\begin{bmatrix}
\mathcal{H}_L \!\left(u^{\mathrm{d}}\right) \\
\mathcal{H}_L \!\left(y^{\mathrm{d}}\right)
\end{bmatrix}
g.
\end{equation}

A prerequisite for invoking the lemma is that the input sequence used for offline data collection must be persistently exciting (PE). An input $u^\mathrm{d}$ is said to be PE of order $L$ if its corresponding Hankel matrix $\mathcal{H}_L(u^\mathrm{d})$ has full row rank. To satisfy this condition in practice, a rich variety of input signals were applied to the soft robotic arm during the offline phase, ensuring that the arm explored a comprehensive portion of its workspace and that the resulting data matrix fully captured its dynamics.

According to the lemma, for any controllable soft robotic arm system driven by a PE input $u^\mathrm{d}$, any feasible motion trajectory of length $L$ can be accurately represented, and any future motion that the robot can achieve can be represented as a linear combination of the behaviors it has exhibited in the past. Building on this principle, DeePC partitions the Hankel matrices into past and future components, which are then used to formulate a constrained optimal control problem. Specifically, let $T_{\mathrm{ini}}$, $N\in \mathbb{Z}$, and $L=T_{\mathrm{ini}}+N$. The following block matrices are defined:
\begin{equation}
\begin{bmatrix}
U_p \\ U_f
\end{bmatrix}
:= \mathcal{H}_L \!\left( u^{\mathrm{d}} \right), 
\quad
\begin{bmatrix}
Y_p \\ Y_f
\end{bmatrix}
:= \mathcal{H}_L \!\left( y^{\mathrm{d}} \right),
\end{equation}
where \( (U_p, Y_p) \) represent the past control inputs and corresponding motion trajectories of the soft robotic arm, comprising the first \( T_{\mathrm{ini}} \) block rows of \( \mathcal{H}_L ( u^{\mathrm{d}} ) \) and \( \mathcal{H}_L ( y^{\mathrm{d}} ) \), respectively. These data are used to connect the model's predictions to the most recently observed behavior of the robotic arm. \( (U_f, Y_f) \) denote the future control inputs and their associated motion trajectories of the soft arm, encompassing the last \( N \) block rows of \( \mathcal{H}_L ( u^{\mathrm{d}} ) \) and \( \mathcal{H}_L ( y^{\mathrm{d}} ) \), thereby forming a predictive basis for all feasible system evolutions. This partitioning provides the basis for casting DeePC as a finite-horizon, data-driven optimal control problem without requiring explicit system identification. 

To achieve precise trajectory tracking of the 3D soft robotic arm, the control task is formulated as a receding-horizon optimization problem within the DeePC framework. With an initial window length $T_{\mathrm{ini}}$ and prediction horizon $N$, the resulting optimization problem can be formulated as:
\begin{equation}
\begin{aligned}
\min_{g,u,y} \quad & \| y - y_r \|_Q^2 + \| u \|_R^2 \\
\text{subject to} \quad & 
\begin{bmatrix}
U_p \\ U_f \\ Y_p \\ Y_f
\end{bmatrix} g =
\begin{bmatrix}
u_{\mathrm{ini}} \\ u \\ y_{\mathrm{ini}} \\ y
\end{bmatrix}, \\
& u \in \mathcal{U}, \; y \in \mathcal{Y},
\end{aligned}
\end{equation}
where $y_r$ is the reference trajectory, $Q$, $R$ are weighting matrices,
$u_{\mathrm{ini}} = \begin{bmatrix}
    u(t-T_{\mathrm{ini}})^{\top}, u(t-T_{\mathrm{ini}}+1)^{\top}, \ldots, u(t-1)^{\top}
\end{bmatrix}^{\top}$ and $y_{\mathrm{ini}} = \begin{bmatrix}
    y(t-T_{\mathrm{ini}})^{\top}, y(t-T_{\mathrm{ini}}+1)^{\top}, \ldots, y(t-1)^{\top}
\end{bmatrix}^{\top}$ are the input and output sequences within a past time horizon of length $T_{\mathrm{ini}}$, $u = \begin{bmatrix}
    u(t)^{\top}, u(t+1)^{\top}, \ldots, u(t+N-1)^{\top}
\end{bmatrix}^{\top}$ and $y = \begin{bmatrix}
    y(t)^{\top}, y(t+1)^{\top}, \ldots, y(t+N-1)^{\top}
\end{bmatrix}^{\top}$ are the input and output sequences within a prediction horizon of length $N$,
and $\mathcal{U}$, $\mathcal{Y}$ represent the admissible sets of inputs and outputs.

\subsection{Robust DeePC via Regularization}
Although DeePC and MPC are equivalent under ideal conditions (i.e., a controllable LTI system with a persistently exciting input) \cite{coulson2019data}, this equivalence rarely holds for the real-world soft robotic arm. The ideal assumptions are challenged by multiple factors: (i) measurement noise from sensors and external disturbances (e.g., airflow, vibrations), and (ii) the highly nonlinear dynamics inherent to soft materials (e.g., viscoelasticity, hysteresis), which cannot be fully captured by a finite dataset. These imperfections mean that a newly observed trajectory $(u_{\mathrm{ini}}, y_{\mathrm{ini}} )$ may not be perfectly linearly represented in the historical database $(U_p ,Y_p)$. This mismatch makes the equality constraint in the optimization problem impossible to satisfy, leading to infeasibility. To restore feasibility and improve robustness against these real-world effects, the standard DeePC formulation is often augmented with regularization. Common approaches include:

\textbf{Slack Variables:} Introduce an output slack variable $\sigma_y$ to relax the requirement of perfectly matching past trajectories. It allows for a small deviation $\sigma_y$ between the soft robotic arm's actual historical trajectory and the best trajectory represented by the data. The deviation is penalized in the cost function, and assigning a sufficiently large weight $\lambda_y$ ensures that $\sigma_y$ becomes non-zero only when the original constraints are infeasible.

\textbf{Regularization on $g$:} Add a penalty term, such as $\lambda_g \| g \|_2^2$, to the cost function. This discourages overfitting to noisy data and improves the generalization and stability of the solution. Physically, this prevents the controller from reacting aggressively to sensor noise, resulting in smoother control actions and reduced jitter in the soft robotic arm's end-effector motion.

The resulting regularized DeePC problem for the soft robotic arm can thus be formulated as:
\begin{equation} \label{equ:deepc}
\begin{aligned}
    \min_{g,u,y,\sigma_y} \quad 
    & \|y - y_r\|_Q^2 + \|u\|_R^2 
      + \lambda_y \|\sigma_y\|_2^2 + \lambda_g \|g\|_2^2 \\
    \text{subject to} \quad 
    & \begin{bmatrix}
        U_p \\ U_f \\ Y_p \\ Y_f
      \end{bmatrix} g
      =
      \begin{bmatrix}
        u_{\text{ini}} \\ u \\ y_{\text{ini}} \\ y
      \end{bmatrix}
      +
      \begin{bmatrix}
        0 \\ 0 \\ \sigma_y \\ 0
      \end{bmatrix}, \\
    & u \in \mathcal{U}, \; y \in \mathcal{Y}.
\end{aligned}
\end{equation}

\textbf{Dimension Reduction:} The complexity of soft robots necessitates extensive offline data acquisition to capture their dynamics, resulting in very large Hankel matrices for the DeePC controller. Consequently, the dimension of the optimization variable $g$ becomes extremely large, making the optimization problem computationally intractable for real-time control loops that require millisecond-level response times. To make DeePC practical for real-time implementation, an effective data-driven dimensionality reduction technique is essential. SVD can be applied to reduce the dimension of the Hankel matrix in~\eqref{equ:deepc}, and thus improve the computational efficiency of the DeePC problem~\cite{zhang2023CSL}.

To implement this, the concatenated Hankel matrix of the collected input–output data is factorized using SVD. Specifically, the decomposition is given by:
\begin{equation} \label{equ:svd}
\begin{bmatrix}
\mathcal{H}_L(u^{\mathrm{d}}) \\
\mathcal{H}_L(y^{\mathrm{d}})
\end{bmatrix}
=
\underbrace{\begin{bmatrix} W_1 & W_2 \end{bmatrix}}_{W}
\underbrace{\begin{bmatrix} \Sigma_1 & 0 \\ 0 & \Sigma_2 \end{bmatrix}}_{\Sigma}
\underbrace{\begin{bmatrix} V_1 & V_2 \end{bmatrix}}_{V^\top}.
\end{equation}
In~\eqref{equ:svd}, $W \in \mathbb{R}^{q_1 \times q_1}$ with $q_{1} = (m+p)L$ and $V \in \mathbb{R}^{q_2 \times q_2}$ with $q_{2} = T-L+1$ are orthogonal matrices. $\Sigma \in \mathbb{R}^{q_1 \times q_2}$ is a rectangular diagonal matrix with the singular values arranged in descending order. $\Sigma_{1} \in \mathbb{R}^{r \times r}$ contains the top $r$ non-zero singular values, and $W_{1}$, $W_{2}$, $\Sigma_{2}$, $V_{1}$, and $V_{2}$ are defined with compatible dimension. 

By truncating this decomposition to the top $r$ singular values, which capture the system's principal dynamics, a new condensed data matrix $\bar{\mathcal{H}}_{L} \in \mathbb{R}^{q_1 \times r}$ is obtained:
\begin{equation}
\bar{\mathcal{H}}_{L} = \mathcal{H}_{L} V_{1} = W_{1} \Sigma_{1}.
\end{equation}

Consequently, the original DeePC optimization problem can be reformulated using the compressed data matrix and a reduced optimization variable $\bar{g} \in \mathbb{R}^r$. The resulting regularized SVD-DeePC problem for the soft arm is given by:
\begin{equation}
\begin{aligned}
\min_{\bar{g},u,y,\sigma_y} \quad & 
\| y - y_r \|_Q^2 + \| u \|_R^2 
+ \lambda_y \| \sigma_y \|_2^2 
+ \lambda_g \| \bar{g} \|_2^2 \\
\text{subject to} \quad & 
\bar{\mathcal{H}}_{L} \bar{g} =
\begin{bmatrix}
u_{\mathrm{ini}} \\ u \\ y_{\mathrm{ini}} \\ y
\end{bmatrix}
+
\begin{bmatrix}
0 \\ 0 \\ \sigma_y \\ 0
\end{bmatrix}, \quad
u \in \mathcal{U}, \; y \in \mathcal{Y}.
\end{aligned}
\end{equation}

This formulation provides a significant computational advantage: the dimension of the optimization variable is reduced from the dataset length $(T-L+1)$ to the approximation rank $r$. Such a reduction makes the approach more suitable for real-time control of the soft robotic arm~\cite{wang2024mechanical}. The choice of $r$ can be guided by inspecting the decay of the singular values or determined based on control performance.

\section{EXPERIMENTS}
This section presents an experimental study aimed at evaluating the performance of DeePC on a soft robotic arm in two representative 3D tasks: (i) fixed-point reaching and (ii) trajectory tracking.

\subsection{Setup and data collection}
\begin{figure}[htbp]
    \centering
    \includegraphics[width=0.32\textwidth]{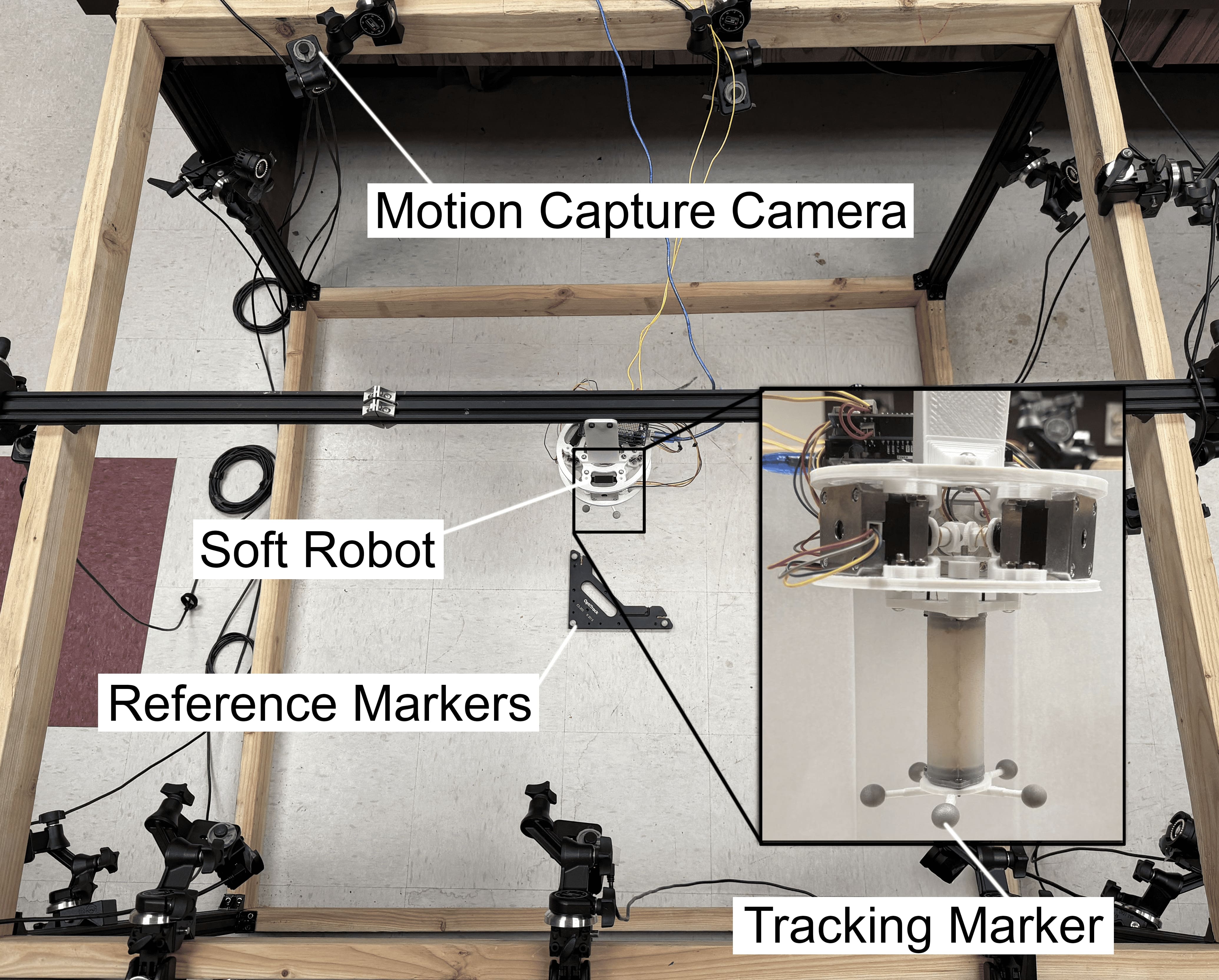}
    \caption{Experimental setup with motion capture cameras and a test frame.}
    \label{fig:exp_setup}
    \vspace{-14pt}
\end{figure}

We validated the proposed method on a self-developed cable-driven soft robotic arm. The experimental setup is shown in Fig.~\ref{fig:exp_setup}. The arm is actuated by three cables, with a total effective length of 90 mm. End-effector positions were tracked in real time using a 12-camera OptiTrack FLEX13 motion capture system. Data collection and control were executed on a host PC, with the DeePC framework implemented in MATLAB. Optimal control commands were transmitted via serial communication to a low-level controller, which drove the stepper motors to adjust cable lengths.

To construct a nonparametric representation, input–output data were collected. Here, $u^\mathrm{d}$ denotes the incremental length of the three actuation cables, while $y^\mathrm{d} = (p_x, p_y, p_z)$ represents the end-effector position in task space. During data acquisition, ramp-and-hold excitations were applied to the low-level controller, and the corresponding end-effector positions were continuously recorded. The resulting dataset was then arranged into a Hankel matrix as described in Section~\ref{sec:deepc}. When sufficiently rich, this data supports an accurate prediction of system behavior.

\subsection{Fixed-point reaching control}
To evaluate DeePC’s ability to accurately drive the soft arm to specified target positions, we conducted a multi-stage reference tracking experiment. The bending angle $\phi_b$ and orientation angle $\gamma_g$ were commanded to follow stepwise reference trajectories across three stages:
(I) $\phi_b = 20^\circ,\ \gamma_g = 0^\circ$; (II) $\phi_b = 40^\circ,\ \gamma_g = 60^\circ$; (III) $\phi_b = 60^\circ,\ \gamma_g = 120^\circ$.
Here, $\phi_b$ and $\gamma_g$ are computed from the tip position using the constant curvature model \cite{hasanshahi2024design}.

\begin{figure}[htbp]
    \centering
    \includegraphics[width=0.45\textwidth]{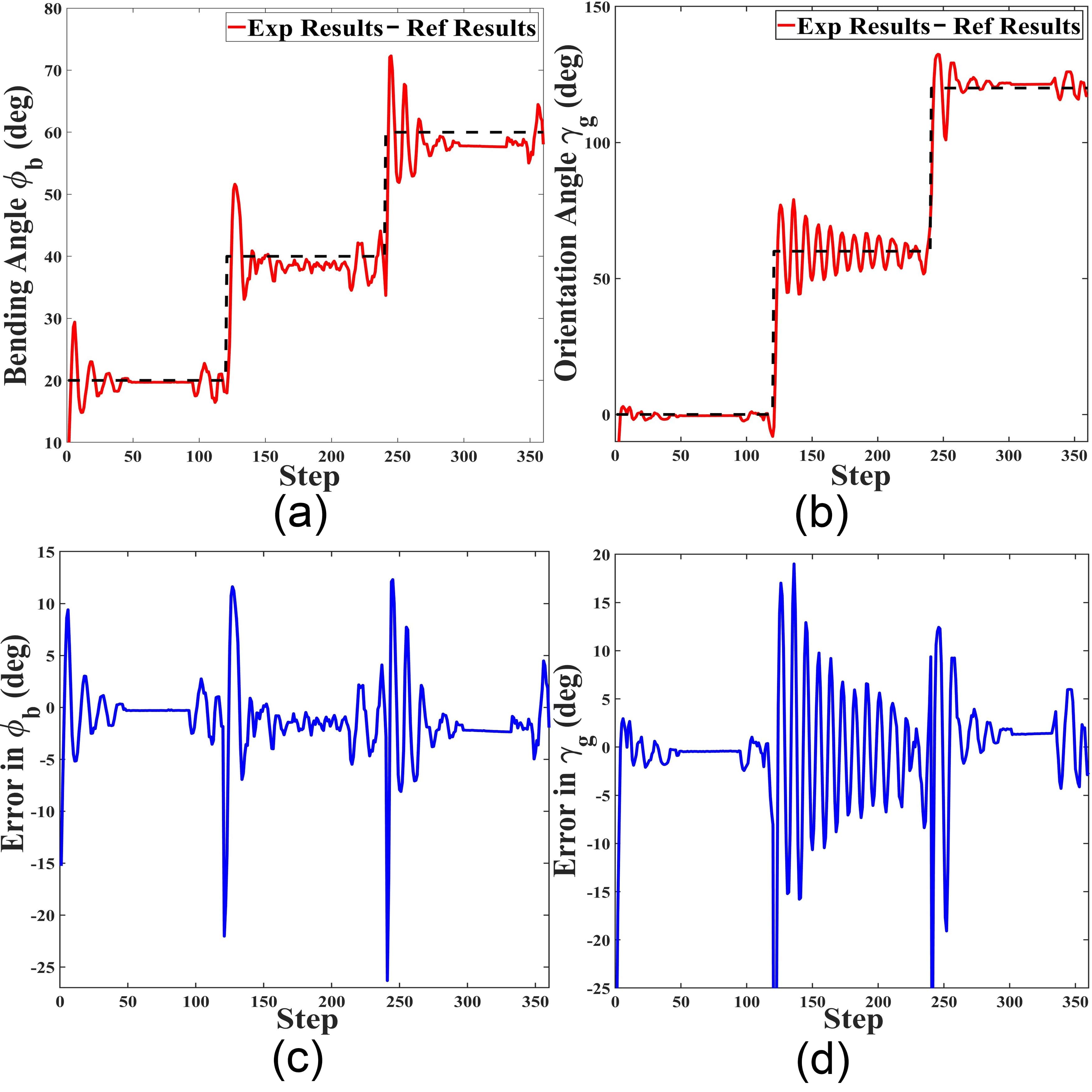}
    \caption{Experimental results of DeePC control. (a–b) Tracking performance of bending angle $\phi_b$ and orientation angle $\gamma_g$. (c–d) Corresponding tracking errors.}
    \label{fig:exp_result1}
    \vspace{-12pt}
\end{figure}

As shown in Fig.~\ref{fig:exp_result1}, DeePC successfully drove the arm to track both the bending angle $\phi_b$ and orientation angle $\gamma_g$. The bending angle converged rapidly in each stage with small steady-state error, exhibiting only limited overshoot and short oscillations before stabilization. For the orientation angle, noticeable oscillations were observed around $60^\circ$, but their amplitude decayed gradually and converged to the reference. In contrast, the $0^\circ$ and $120^\circ$ stages displayed relatively mild transients and fast settling. Overall, the results demonstrate DeePC’s effectiveness in achieving accurate target reaching and stage switching without requiring an explicit model. Further improvements may be obtained by tuning the weighting and regularization parameters to suppress residual overshoot and oscillations.

\begin{figure}[]
    \centering
    \begin{subfigure}{0.25\textwidth}
        \centering
        \includegraphics[width=\linewidth]{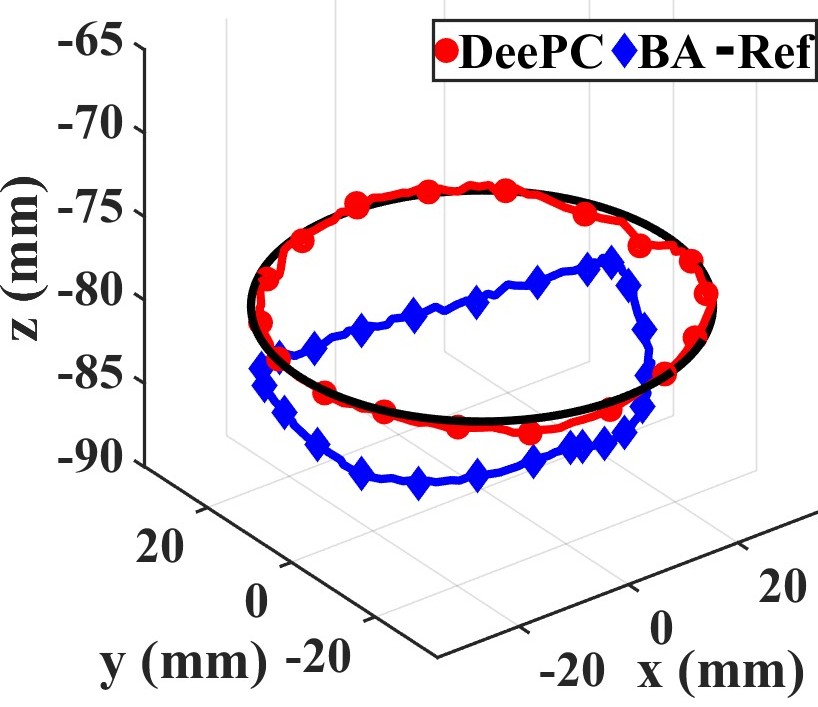}
        \caption{}
        \label{exp21}
    \end{subfigure}
    \hfill
    \begin{subfigure}{0.229\textwidth}
        \centering
        \includegraphics[width=\linewidth]{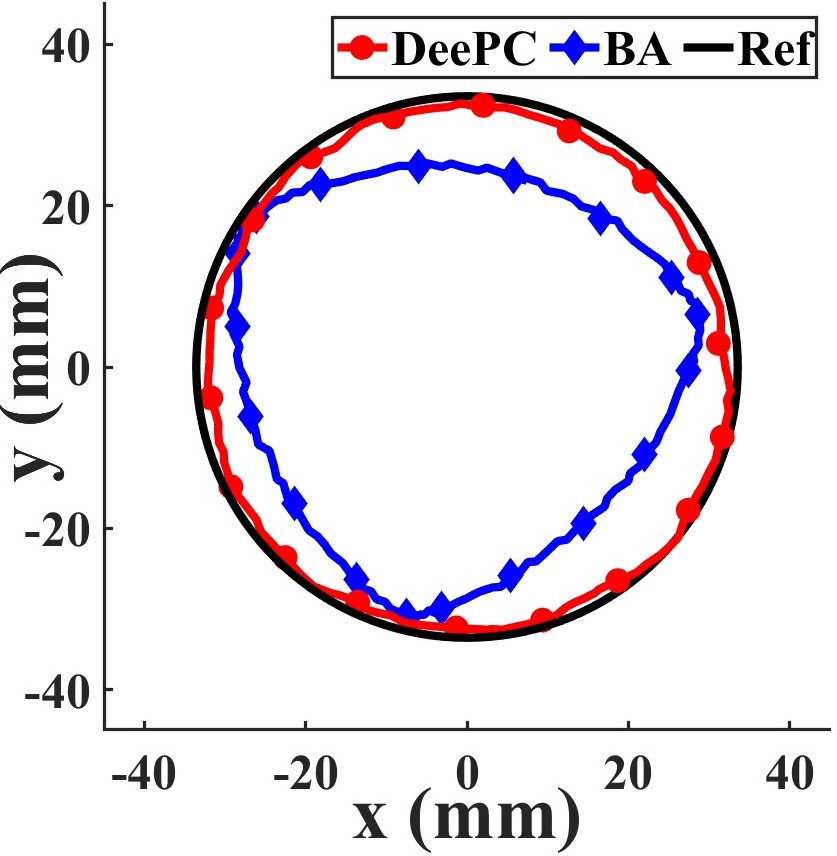}
        \caption{}
        \label{exp22}
    \end{subfigure}
    \caption{Trajectory tracking control results. (a) Comparison of 3D tracking performance. (b) Comparison of 2D tracking performance in the $x$–$y$ plane.}
    \label{fig:exp_results2}
    \vspace{-14pt}
\end{figure}

\subsection{Trajectory tracking control}
In this experiment, the arm was tasked to track a circular trajectory in task space. The reference trajectory was defined as a circle of radius $r$, discretized into a sequence of waypoints. For the baseline, control inputs were computed directly from the geometric model. For DeePC, the control inputs were constrained to $u \in [0,90]$, with $T_{\mathrm{ini}} = 20$ and $N = 30$. The weighting matrices were set to $Q = 10 \cdot I_{3 \times 3}$ and $R = 2 \times 10^{-3} \cdot I_{3 \times 3}$, while the regularization parameters were chosen as $\lambda_g = 300$ and $\lambda_y = 1,000$.

For comparison, we implemented a baseline model-based controller derived from the constant-curvature assumption \cite{qi2024design}. In this framework, the relationship between the backbone curvature of the arm and the cable length is expressed through idealized geometric relations. Specifically, let $\kappa_{c,i}$ and $R_{c,i}$ denote the curvature and radius of curvature of the $i$-th cable, respectively, and let $l_i$ denote its length. For the $i$-th cable in a section:
\begin{equation}
\frac{1}{\kappa_b} = \frac{1}{\kappa_{c,i}} + d_i
\end{equation}
\begin{equation}
l_i = R_{c,i} \cdot \phi_b 
= \frac{\kappa_b}{\kappa_{c,i}} L
\end{equation}
where $d_i$ represents the distance between the incident point of the $i$-th cable and the neutral plane, $\kappa_b$ is the backbone curvature, and $L$ is the segment length.

\begin{figure*}[ht!]
    \centering
    \includegraphics[width=0.8\textwidth]{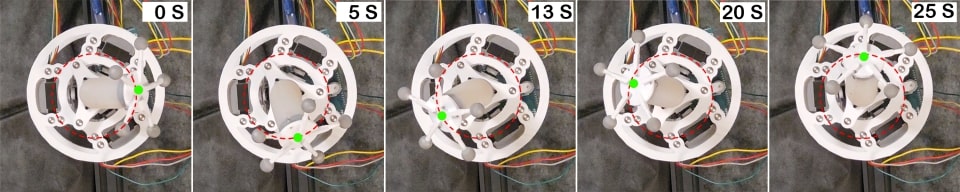}
    \caption{Movement and bending configurations of the soft arm during circular trajectory tracking.}
    \label{fig:MBC}
    \vspace{-14pt}
\end{figure*}
As shown in Fig.~\ref{fig:exp_results2}, DeePC achieved precise tracking with minimal deviation from the reference trajectory, both in 3D space (Fig.~\ref{exp21}) and in the $x$–$y$ projection (Fig.~\ref{exp22}). The DeePC trajectory closely overlapped the reference circle, maintaining a consistent radius and smooth curvature. In contrast, the baseline approach produced a distorted path: in 3D, the baseline trajectory sagged below the desired plane, while in the $x$–$y$ projection the circle was compressed into an irregular, triangular-like shape. These distortions reflect the limitations of constant-curvature geometric assumptions, which cannot capture the soft arm’s nonlinear bending and torsional behaviors.
The physical bending configurations of the soft arm during circular trajectory tracking are illustrated in Fig.~\ref{fig:MBC}. The sequential snapshots show that DeePC successfully generated smooth, coordinated cable motions, enabling the arm to trace the desired circular path in real time.

\vspace{-3pt}
\section{CONCLUSION}
\vspace{-5pt}
This paper addressed the challenge of high-performance, three-dimensional dynamic control for soft robots. We designed, implemented, and experimentally validated a data-enabled predictive control (DeePC) framework on an optimized cable-driven soft arm platform. The experiments confirmed the effectiveness of DeePC for precise 3D soft robot control, advancing beyond prior work limited to planar systems or open-loop control, and demonstrated its ability to handle the complex, nonlinear dynamics of soft robots without requiring explicit first-principles models.
Future work will focus on evaluating performance under large external disturbances and varying payloads. Another important direction is extending this framework from single-segment to multi-segment soft robotic arms, enabling more dexterous and versatile manipulation in 3D space. Furthermore, applying the method to advanced tasks, such as fruit handling and object manipulation, represents a promising step toward expanding the practical capabilities of soft robotic systems.

\vspace{-3pt}
\bibliography{ref}
\bibliographystyle{IEEEtran}

\end{document}